\pdfoutput=1
\documentclass[11pt]{article}

\usepackage[]{acl}

\usepackage{times}
\usepackage{latexsym}

\usepackage{hyperref}
\usepackage{url}
\usepackage{multirow}
\usepackage{graphicx}
\usepackage{comment}
\usepackage{amssymb}
\usepackage{amsmath}
\usepackage{url}

\usepackage[T1]{fontenc}
\usepackage[utf8]{inputenc}

\usepackage{microtype}

\title{Uncertainty Estimation on Sequential
Labeling\\ via Uncertainty Transmission}

\author{
Jianfeng He{$^\dag$}, Linlin Yu{$^\ddag$}, Shuo Lei{$^\dag$},
Chang-Tien Lu{$^\dag$}, Feng Chen{$^\ddag$}
\\ 
 {$^\dag$}Department of Computer Science, Virginia Tech, Falls Church, VA, USA\\
{$^\ddag$}Department of Computer Science, The University of Texas at Dallas, Richardson, TX, USA\\
 {$^\dag$}\{jianfenghe, slei, ctlu\}@vt.edu, \\{$^\ddag$}\{Feng.Chen, Linlin.Yu\}@utdallas.edu
 }

\begin{document}
\maketitle

\begin{abstract}
Sequential labeling is a task predicting labels for each token in a sequence, such as Named Entity Recognition (NER). NER tasks aim to extract entities and predict their labels given a text, which is important in information extraction. Although previous works have shown great progress in improving NER performance, uncertainty estimation on NER (UE-NER) is still underexplored but essential. This work focuses on UE-NER, which aims to estimate uncertainty scores for the NER predictions. Previous uncertainty estimation models often overlook two unique characteristics of NER: the connection between entities (i.e., one entity embedding is learned based on the other ones) and wrong span cases in the entity extraction subtask. Therefore, we propose a Sequential Labeling Posterior Network (SLPN) to estimate uncertainty scores for the extracted entities, considering uncertainty transmitted from other tokens. Moreover, we have defined an evaluation strategy to address the specificity of wrong-span cases. Our SLPN has achieved significant improvements on three datasets, such as a 5.54-point improvement in AUPR on the MIT-Restaurant dataset. Our code is available at \url{https://github.com/he159ok/UncSeqLabeling_SLPN}.

\end{abstract}

\section{Introduction}
Named entity recognition (NER) is a popular task in the information extraction domain~\cite{lample2016neural}, which involves two steps, detecting entity spans and predicting the entity labels. 
In many information extraction scenarios, there are significant consequences for relying on inaccurate NER predictions. For example, extracting an inaccurate time can lead to erroneous policy analysis, or misclassifying a person's name for a time can result in a privacy breach. Therefore, it is crucial to determine whether we can trust the NER predictions or not.
As a result, our goal is to enhance Uncertainty Estimation in NER (UE-NER), which aims to quantify prediction confidence in NER tasks. 

\begin{figure*}[!htb]
\centering
\includegraphics[width=0.95\textwidth]{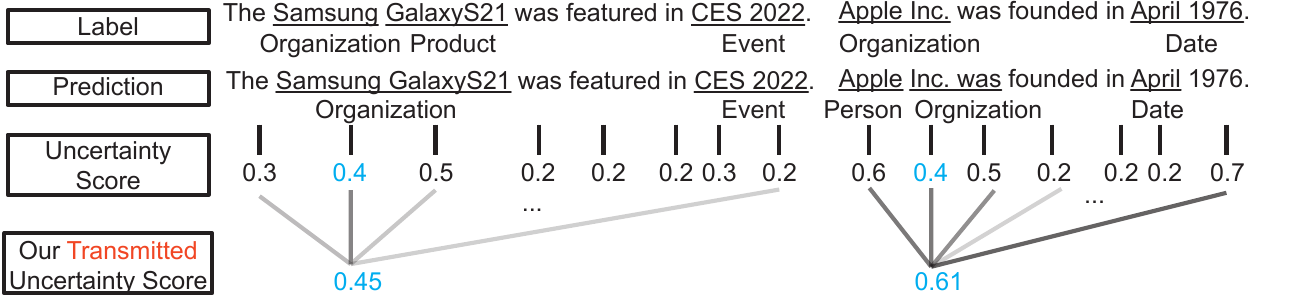}
\caption{
In this example, though the tokens ``Samsung'' and ``Inc.'' both have the same uncertainty score of 0.4, the context in the right case exhibits higher uncertainty. This suggests that ``Inc.'' should be considered more uncertain than ``Samsung.'' Therefore, we propose transmitting the predicted uncertainty from other tokens to a given token.
}
\label{fig:compare}
\end{figure*}

The NER task differs from general classification (e.g., text classification~\cite{minaee2021deep} in two key ways, making previous uncertainty estimation models suboptimal for UE-NER.

First, the predicted entity labels in the NER task are directly dependent on the token embeddings, and uncertainty transmission between token embeddings is unique in NER. Concretely, given an example text ``\underline{Barack Obama} was born in Honolulu, Hawaii,'' the entity label ``person'' applies to ``Barack Obama.'' The embedding of the token ``Obama'' is obtained by accumulating its own embedding and embeddings from other tokens in Recurrent Neural Network~\cite{medsker2001recurrent} and transformer~\cite{vaswani2017attention}. 
Consequently, if a token embedding has higher uncertainty, the other token embedding will have more transmitted uncertainty from the token. Since token embeddings directly affect token labels and further affect entity labels, high uncertainty in a token embedding will result in a predicted entity label with high uncertainty. Therefore, in the context of UE-NER, a token uncertainty in UE-NER consists the individual token uncertainty and the uncertainty transmitted from other tokens.

However, the current uncertainty estimation methods ignore the uncertainty transmission between tokens.
Especially, current uncertainty estimation methods can be classified into two main categories: parameter-distribution-based methods, such as Bayesian Neural Networks (BNN)\cite{osawa2019practical, maddox2019simple}, which learns a distribution over the model parameters; and sample-distribution-based methods, which calculate uncertainty scores based on the distribution of training samples~\cite{charpentier2020posterior,he2020towards,park2018adversarial}.
These methods primarily focus on image or text classification, where correlations between different images or texts are weak or limited. Consequently, they overlook the uncertainty transmission inherent in sequential labeling.
Since sequential labeling plays a pivotal role in Natural Language Processing (NLP), with NER as a representative example, it is imperative for us to address UE-NER by considering uncertainty transmission, shown as Figure~\ref{fig:compare}.

The second  characteristic of NER tasks is that they involve an additional step, entity extraction, besides entity classification.
In contrast to previous text classification tasks~\cite{minaee2021deep}, which focus solely on sample classification, NER tasks require the additional task of extracting entity spans, such as locating ``Barack Obama.''
However, entity span extraction may predict entities with wrong span (WS), such as predicting ``Obama was'' as an entity. These WS entities lack ground truth entity labels and evaluating uncertainty estimation requires ground truth labels, thus these entities cannot be used for evaluating uncertainty estimation. 
Therefore, we require an innovative approach to evaluate a UE-NER model that takes into account these WS entities.

To address the first issue, we propose a Sequential Labeling Posterior Network (SLPN) for transmitting uncertainty. This network is built upon an evidential neural network framework~\cite{charpentier2020posterior} with a novel design to transmit uncertainty from other tokens. For the second issue, we categorize the ground truth entities and predicted entities into three groups: unique entities in the ground truth, unique entities in the prediction, and shared entities between the ground truth and prediction. 
We, then, treat WS entity detection as a separate subtask, in addition to out-of-domain (OOD) detection, which is a common task used to evaluate uncertainty estimation~\cite{zhao2020uncertainty}. 
The WS and OOD detections use different combinations of the three-group entities. 
Furthermore, we evaluate the performance of a UE-NER model by computing a weighted sum of WS entity detection and OOD detection performance, providing a comprehensive assessment of the UE-NER model.
Our contributions are as follows.

\begin{itemize}    
    \item Since each token embedding is influenced by other tokens within a given text, and token embedding directly affects the uncertainty of predicted entity labels, we propose a novel method to transmit uncertainty between tokens using a revised self-attention. To the best of our knowledge, we are the first to consider uncertainty transmission in UE-NER.
    
    \item 
    Because of the existence of WS entities in the NER task, we have found that traditional evaluation methods for uncertainty estimation are inapplicable in UE-NER. Therefore, we propose a novel uncertainty estimation evaluation to evaluate both OOD and WS detection tasks.

\end{itemize}

\section{Related Work}
\noindent\textbf{Named Entity Recognition.}
Named Entity Recognition (NER) is a task focused on extracting and classifying entities within text. It serves as a prominent example of sequential labeling, where each token in a sequence is assigned a label. Various techniques have been employed for NER, including Recursive Neural Networks (e.g., LSTM~\cite{hammerton2003named}), pretrained transformer models (e.g., BERT~\cite{devlin2018bert}). In some cases, Conditional Random Fields (CRF) are incorporated into token encoders, such as LSTM+CRF~\cite{lample2016neural}, to enhance performance.

Further, recent experiments have explored the use of Large Language Models (LLMs) for NER. An LLM-based approach treats NER as a generative task, with each turn generating one category of entities~\cite{wang2023gpt}. However, it is noticeable that~\citet {wang2023gpt} found that GPT3-based NER solutions~\cite{floridi2020gpt} did not outperform pretrained transformer based method. 
Since both pretrained transformer-based methods and LLMs are built on transformer architectures~\cite{vaswani2017attention} and pretrained transformer-based methods take NER as sequential generation rather than sequential labeling, as well as perform better than GPT-3 on the NER task, our research focuses on UE-NER using pretrained-transformer-based methods.

\noindent\textbf{Uncertainty estimation on natural language processing.} Generally, for the usage of uncertainty estimation on training data, the uncertainty score helps with sample selection in active learning~\cite{wang2021meta}. For usage on the testing data, uncertainty estimation mainly serves two tasks: \textit{OOD detection}~\cite{hart2023improvements}, where the testing samples include OOD samples, and the task aims to identify these OOD samples; and ~\textit{misclassified result detection}: where testing samples are in-domain~\cite{zhang2019mitigating,he2020towards,hu2021uncertainty}. Our work specifically focuses on OOD detection in the testing samples.

In the NER domain,~\citet{nguyen2021loss,chang2020using,liu2022ltp} estimated uncertainty scores on unlabeled training data for active learning.~\citet{vazhentsev2022uncertainty} were the first to apply uncertainty estimation to address misclassification in NER testing data using techniques like dropout~\cite{gal2016dropout} and deterministic uncertainty estimation methods (e.g., Gaussian process ~\cite{NEURIPS2020_543e8374}). Additionally, on the testing samples,~\citet{zhang2023ner} were the first to apply uncertainty estimation to detect OOD instances in NER testing data. Compared to~\citet{zhang2023ner}, who assigned different weights to different tokens, our work focuses on the transmission of uncertainty from other tokens to a specific token.

\section{UE-NER Task Setting}
Before we introduce UE-NER, we first introduce NER tasks, which is a representative sequential labeling task. Given a text $\mathbf{X}=[\mathbf{x}_1, \mathbf{x}_2, ..., \mathbf{x}_n]$ with $n$ tokens, where $\mathbf{x}_i\in \mathbb{R}^{h\times 1}$ is an embedding of a token, NER task aims at learning a NER model predicting their token labels. 
Then, the entities are extracted by the token labels based on the BIOES mechanism~\cite{chiu2016named} (e.g., ``Brack'' with B-PER label, and ``Obama'' with I-PER label). Moreover, the extracted entities are classified by merging the entity tokens. For example, ``Brack Obama'' is categorized as a Person because these two tokens are categorized as the beginning and intermediate of the person label.

For the UE-NER task, we aim to learn a UE-NER model $\Phi$ to predict the confidence of each predicted token label. 
We apply $\Phi$ for OOD detection, which is a common task to evaluate uncertainty estimation~\cite{zhao2020uncertainty}.
Concretely, the training data and validation data for $\Phi$ are the in-domain (ID) text without OOD entities. 
The testing data of $\Phi$ includes both ID text and OOD text, where OOD text has both ID and OOD entities. 
A better $\Phi$ should detect more OOD entities in the testing set and have better NER performance.

\section{Preliminary: Posterior Network}
The parameter-distribution-based 
uncertainty estimation method is usually implemented via ensemble sampling~\cite{gal2016dropout} and thus requires multiple forward passes to estimate uncertainty, which is time-consuming. In contrast, Evidential Deep Learning (EDL)~\cite{sensoy2018evidential} is a representative sample-distribution-based 
uncertainty estimation method and is implemented via a deterministic model, thus requiring only one forward pass to estimate uncertainty. Due to its efficiency, we choose EDL.

In EDL, considering the classification task and given the input vector $\mathbf{X} $, the class prediction $y \in [c]$ 
for an input sample follows a categorical distribution with $c$ classes.The categorical distribution naturally follows a Dirichlet distribution, i.e.
\begin{equation}
    y\sim \text{Cat}(\mathbf{p}) ,  \ \  \mathbf{p} \sim \text{Dir}(\boldsymbol{\alpha})
\end{equation}

The expected class probability $\bar{\mathbf{p}}$ is calculated as below,
\begin{equation}
\alpha_0 = \sum_{k=1}^{c} \alpha_k , \ \ 
\bar{\mathbf{p}} = \frac{\boldsymbol{\alpha}}{\alpha_0} 
\label{eq:sum_alpha}
\end{equation}

where $\text{Dir}(\boldsymbol{\alpha})$ is an approximation of the posterior distribution of class probabilities, conditioned on the input feature vector. 
The concentrate parameters $\boldsymbol{\alpha}=[\alpha_1, \alpha_2, ..., \alpha_{c}]$ can be interpreted as the evidence for the given example belonging to the corresponding class \cite{jsang2018subjective}. The evidence is the count of pseudo support from training samples.

As a representative model of EDL, Posterior Network (PN) \cite{charpentier2020posterior} is originally designed for image classification and involves two main steps. First, a feature encoder 
maps the raw features into a low-dimensional latent space. Second, a normalizing flow such as Radial \cite{rezende2015variational} is used to estimate class-wise density on the latent space, which is proportional to the class-wise evidence. Essentially, a greater density for a particular class implies stronger evidence belonging to this class for the given example. 

PN is trained with the sum of two loss $\mathcal{L}^{\text{UCE}}$ and $\mathcal{L}^{\text{ER}}$ for $N$ training samples as below,
\begin{equation}
\label{eq:ori_uce}
    \mathcal{L}^{\text{UCE}} = \frac{1}{N}\sum_{i=1}^N\mathbb{E}_{\mathbf{p}_i \sim \text{Dir}(\mathbf{p}_i|\boldsymbol{\alpha}_i)}[\text{CE}(\mathbf{p}_i, y_i)]
\end{equation}
\begin{equation}
    \mathcal{L}^{\text{ER}} = - \frac{1}{N}\sum_{i=1}^N \mathbb{H}(\text{Dir}(\mathbf{p}_i|\boldsymbol{\alpha}_i))
\end{equation}
where the Uncertainty Cross Entropy (UCE) loss $\mathcal{L}^{\text{UCE}}$ encourages high evidence for the ground-truth category
and entropy regularization $\mathcal{L}^{\text{ER}}$ encourages a smooth Dirichlet distribution.

\section{Model}
\begin{figure*}[htb]
\centering
\includegraphics[width=1.0\textwidth]{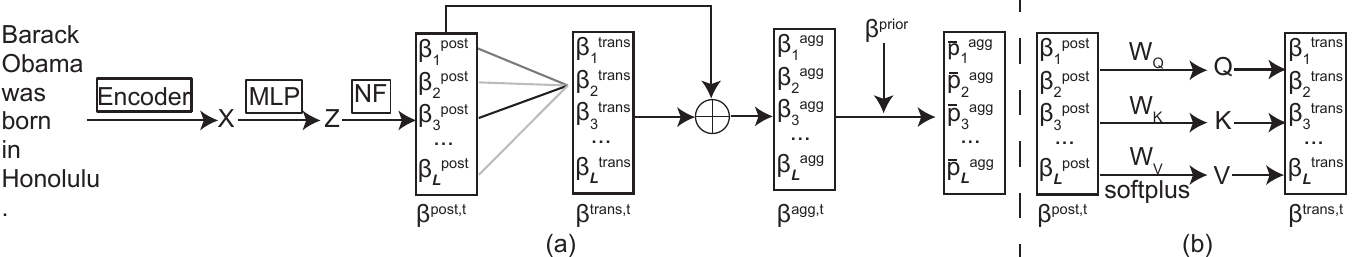}
\caption{
(a) A diagram of our SLPN model illustrates how we achieve uncertainty transmission through a revised self-attention mechanism applied to all tokens. Specifically, the SLPN model begins by generating a text embedding matrix $\mathbf{X}$ with $l$ rows, corresponding to a text containing $l$ tokens. Next, an MLP model projects $\mathbf{X}$ into a latent embedding matrix $\mathbf{Z}$ also with $l$ rows. This $\mathbf{Z}$ matrix is used to compute $\boldsymbol{\beta}^{post,t}\in\mathbb{R}^{l \times c}$ through a normalizing flow (NF) operation. Each row of $\boldsymbol{\beta}^{post,t}$ represents the evidence count  from the token's self-view, directly influencing the uncertainty of each token's prediction.
In contrast to previous research, our approach includes the transmission of uncertainty from all tokens within the text to obtain the transmitted uncertainty $\boldsymbol{\beta}^{trans,t}$. Finally, we combine the sum of $\boldsymbol{\beta}^{post,t}$ and $\boldsymbol{\beta}^{trans,t}$ to generate the semantic matrix $\mathbf{\bar{p}^{agg}}\in\mathbb{R}^{l \times c}$, representing the semantics of the $l$ tokens.
(b) Revised self-attention mechanism.
}
\label{fig:SLPN_frame}
\end{figure*}

We choose PN as it does not require OOD data during training. In contrast, Prior Network~\cite{malinin2018predictive}, another representative EDL method, necessitates OOD data in training. Furthermore, even if OOD data is available, it may not cover all possible OOD scenarios. Thus, we opt for uncertainty transmission based on PN.

\subsection{Our Token-Level Posterior Network}
\label{sec:post_net}
The PN, originally for image classification, is applied to NER for the first time to our knowledge. To better apply PN in NER, we first analyze the difference between tokens and samples (e.g., images). Concretely, tokens can be selected from specific sets, allowing for the calculation of token-level categorical distributions. In contrast, for samples, the vast and continuous potential space of unique samples makes it impractical to compute categorical distributions for every possible sample.

To apply PN into UE-NER and consider the above difference, we propose token-level PN, where we propose to calculate a unique categorical distribution for each token, rather than computing a single shared categorical distribution for all samples. This is because each token exhibits distinct semantic characteristics (e.g., ``Paris'' is more likely to represent a location than ``August''), and thus needs individual categorical distributions. Concretely, a  categorical distribution $\text{Cat}(\mathbf{p}_i)$ of $i$-th token in a text is the total occurrence of $i$-th token in each of $c$ classes given a training set. 
For example, the token ``Apple'' in the training data has 200 and 800 occurrences for the organization and food classes respectively, then ``Apple'' has categorical distribution as $[0, ..., 0.2,.., 0.8, 0...]\in\mathbb{R}^{c}$.

Then, since the classification is usually taken as a multinomial distribution, we can represent the classification as a posterior distribution as below,
\begin{equation}
\mathbb{P}(\mathbf{p}_i|{y}_{i})\varpropto\mathbb{P}({y}_{i}|\mathbf{p}_i)\times\mathbb{P}(\mathbf{p}_i)
\label{eq:emission_bayesian}
\end{equation}
we represent its prior distribution by a Dirichlet distribution $\mathbb{P}(\mathbf{p}_i)=\text{Dir}(\boldsymbol{\beta}^{prior})$, where $\boldsymbol{\beta}^{prior}$ is the parameter of the prior Dirichlet distribution. In practice, we set $\boldsymbol{\beta}^{prior}=\mathbf{1}$ for a flat equiprobable prior when the model brings no initial evidence. 
Due to the conjugate prior property, the posterior distribution can also be represented by a Dirichlet distribution: $\mathbb{P}(\mathbf{p}_i|{y}_{i})=\text{Dir}(\boldsymbol{\beta}^{prior}+\boldsymbol{\beta}^{post}_i)$. The $\boldsymbol{\beta}^{post}_{i}$ is taken as the evidence count for $i$-th token.  
To learn ${\boldsymbol{\beta}^{post}_i}$, PN firstly projects $i$-th token embedding $\mathbf{x}_i$ to a low-dimensional latent vector $\mathbf{z}_i=f(\mathbf{x}_i)$. 
Then, PN learns a normalized probability density $\mathbb{P}(\mathbf{z}_i|k;\theta)$ per class on this latent space. PN then counts the evidence for $k$-th class at $\mathbf{z}_i$ as below:
\begin{equation}
\boldsymbol{\beta}_{i,(k)}^{post}=N\times \mathbb{P}(\mathbf{z}_i|k;\theta)\times\mathbb{P}(k_i)
\label{eq:beta_post_single}
\end{equation}
where $\mathbb{P}(k_i)$ is the probablity that $i$-th token belongs to $k$-th class, extracted from $\text{Cat}(\mathbf{p}_i)$. 
And $\boldsymbol{\beta}_{i}^{post}\in\mathbb{R}^{c}=[\beta_{i,(1)}^{post}, \beta_{i,(2)}^{post}, ..., \beta_{i,(c)}^{post}]$. The $\boldsymbol{\beta}_{i}^{post}$ can be understood as the evidence distribution for $i$-th tokens.
For a text with $l$ tokens, we can concatenate all $l$ tokens' evidence distribution vector $\boldsymbol{\beta}^{post}$ and have $\boldsymbol{\beta}^{post, t}\in\mathbb{R}^{l\times c}$.

\noindent\textbf{Difference to original posterior network.}
Compared to the original sample-level posterior network~\cite{charpentier2020posterior}, which operates at the sample level, our token-level PN differs in two key ways: (1) We use a token-level categorical distribution instead of a sample-level categorical distribution shared among all samples. 
% \lin{[Lin: shared categorical distribution?]} 
(2) We concatenate the $\boldsymbol{\beta}^{post}$ values for each of the $l$ tokens to create a new matrix $\boldsymbol{\beta}^{post, t}\in\mathbb{R}^{l\times c}$ to facilitate uncertainty transmission in Sec.~\ref{sec:slpn}, a step not required in the original PN.

\subsection{Our SLPN} 
\label{sec:slpn}
Though the token-level PN counts the evidence given a token, it ignores the relation between tokens. Shown as Fig.~\ref{fig:compare}, imagine that Token A comes from Text A, and Token B comes from Text B. Token A and Token B have the same predicted uncertainty in terms of token label when only considering the token itself. If the other tokens in Text A  have more uncertainty than other tokens in Text B, then in this case, Token A should be more uncertain than Token B due to the impact of other tokens. 
Thus, we propose a Sequential Labeling Posterior Network (SLPN), which takes the uncertainty impact transmitted from other tokens into consideration. 

Concretely, shown as Figure~\ref{fig:SLPN_frame}(a), a token embedding has accumulated all other token embeddings by the Bidirectional RNN~\cite{huang2015bidirectional} or transformer~\cite{vaswani2017attention}. As a result, token uncertainty should comprise two components: uncertainty originating from the token itself and uncertainty transmitted from other tokens. Since the uncertainty in EDL  depends on the evidence count vector $\boldsymbol{\beta}\in\mathbb{R}^{c}$, we can represent the aggregated uncertainty $\boldsymbol{\beta}_{i}^{agg}\in\mathbb{R}^{c}$ for $i$-th token as below, 
\begin{equation}
{\boldsymbol{\beta}_{i}}^{agg}={\boldsymbol{\beta}_{i}^{post}}+{\boldsymbol{\beta}_{i}^{trans}}
\label{eq:beta_post_agg}
\end{equation}
where $\boldsymbol{\beta}_{i}^{post}$ is the uncertainty coming from the token itself and 
$\boldsymbol{\beta}_{i}^{trans}\in\mathbb{R}^{c}$ is the transmitted uncertainty from all tokens to $i$-th token in the text.
The calculation of $\boldsymbol{\beta}_{i}^{post}$ is described in Sec.~\ref{sec:post_net}.

\noindent \textbf{Calculation of impact transmission weight $\boldsymbol{\beta}_{i}^{trans}$ .} 
Since $\boldsymbol{\beta}_{i}^{trans}$ accumulates all the impact from all tokens in a text, we calculate $\boldsymbol{\beta}_{i}^{trans}$ in a way motivated by self-attention~\cite{vaswani2017attention}. Concretely, we have three projector matrices $W_Q\in\mathbb{R}^{c\times p}$, $W_K\in\mathbb{R}^{c\times p}$ and $W_V\in\mathbb{R}^{c\times c}$ to get the query $Q\in\mathbb{R}^{l\times p}$, key $K\in\mathbb{R}^{l\times p}$ and value $V\in\mathbb{R}^{l\times c}$ as below, 
\begin{equation}
\begin{aligned}
&Q=\boldsymbol{\beta}^{post, t}W_Q, K=\boldsymbol{\beta}^{post, t}W_K\\
% &\\
&V={softplus}(\boldsymbol{\beta}^{post, t}W_V)\\
\end{aligned}
\label{eq:QKV}
\end{equation}
where $p$ is a pre-set dimension.
Different from self-attention, we keep the shape of the $V$ the same as $\boldsymbol{\beta}^{post, t}$, because the $\boldsymbol{\beta}^{post, t}$ has the evidence distribution and we want to avoid multiple projections that might lose the evidence distribution. Besides, we apply the ${softplus}$ activation function~\cite{sun2020convolutional} to make sure the value of $V$ is always greater than 0. We require evidence greater than 0 because EDL is an evidence acquisition process where each training sample adds support to learn higher order evidence distribution, 
and thus evidence can only be increased and not decreased~\cite{amini2020deep,wang2023deep}.
Then, we get the transmitted uncertainty $\boldsymbol{\beta}^{trans, t}\in\mathbb{R}^{l \times c}$ as below,
\begin{equation}
\begin{aligned}
\boldsymbol{\beta}^{trans, t}={softmax}(\frac{QK^T}{\gamma})V
\end{aligned}
\label{eq:trans_unc}
\end{equation}

where $\gamma$ is the hyperparameter to rescale the weight to avoid gradient explosion. More explanation is given in Sec.~\ref{sec:exp_softplus}.

\noindent \textbf{Training Loss.}
Once we have obtained $\boldsymbol{\beta}^{agg}$ using Eq.~\ref{eq:beta_post_agg}, we train our SLPN model via below loss.
\begin{equation}
\begin{aligned}    
L=
&\frac{1}{N}\sum_{i=1}^{N}\mathbb{E}_{\mathbf{p}_{i}^{agg} \sim \text{Dir}(\mathbf{p}_{i}^{agg}|\boldsymbol{\alpha}_{i}^{agg})}[\text{CE}(\mathbf{p}_{i}^{agg}, \mathbf{y}_i)] \\
&-\lambda \frac{1}{N}\sum_{i=1}^{N}\mathbb{H}(\text{Dir}(\mathbf{p}_{i}^{agg}|\boldsymbol{\alpha}_{i}^{agg}))
\end{aligned}
\label{eq:slpn_loss}
\end{equation}
where $\boldsymbol{\alpha}_i^{agg} = \boldsymbol{\beta}_i^{agg} +  \boldsymbol{\beta}^{prior}$ and the expected aggregated class probability of the $i$-th token calculated based on $\beta_{j}^{agg}$ is below,
\begin{equation}
\begin{aligned}
&\mathbf{\bar{p}}_{i}^{agg}=\frac{\boldsymbol{\beta}^{agg}_i+\boldsymbol{\beta}^{prior}}{\sum_{k=1}^c(\beta^{agg}_{i,(k)}+\beta^{prior}_{k})}
\end{aligned}
\label{eq:enn_p}
\end{equation}
where $\boldsymbol{\beta}^{prior}\in\mathbb{R}^c$ is the vector with all default values as 1. As a result, the first item in Eq.~\ref{eq:slpn_loss} is the UCE loss in the token level like Eq.~\ref{eq:ori_uce}, and the second item in Eq.~\ref{eq:slpn_loss} is a regularization encouraging a smooth Dirichlet distribution for each token.

\section{Experiments}
\subsection{Experimental Setup}
\begin{table*}[]
\small
\centering
\caption{The table lists the applied entities for OOD and WS tasks. Recall that original ground-truth entities are $e^{og}=e^s+\hat{e}^g$ (used for OOD detection subtask), new ground-truth entities are $e^{ng}=e^s+\hat{e}^g+\hat{e}^{p}$ (used for WS detection subtask). The values in the brackets are the possible ground truth label values. }
\begin{tabular}{l|ccc}
\hline
                           & $e^s$             & $\hat{e}^g$       & $\hat{e}^p$ \\
\hline
Ground-truth entity labels &ID or OOD & ID or OOD & WS          \\
\hline
OOD detection subtask usage          & use (0 or 1)                 & use (0 or 1)                 & do not use   (N/A)        \\
WS detection subtask usage           & use (0)                 & use(0)                & use (1)         \\
\hline
\end{tabular}
\label{tab:ood_intro}
\end{table*}
\subsubsection{Dataset Setup}
\noindent \textbf{Dataset.} We apply three public datasets: (1) MIT-Restaurant (\textbf{MIT-Res}) dataset is in the restaurant domain with a total of 9181 texts with 8 semantic classes, excluding the ``O'' class. (2) Movie-Simple (\textbf{Mov-Sim}) dataset is in the movie domain with a total of 12,218 texts with 12 semantic classes, excluding the ``O'' class. (3) Movie-Complex (\textbf{Mov-Com}) dataset is also in the movie domain with a total of 9769 texts with 12 semantic classes, excluding the ``O'' class. These three datasets are provided in a common NER framework, Flair~\cite{akbik2019flair}. The criteria of the dataset choice are detailed in Sec.~\ref{sec:criteria_dataset}.

\begin{table*}[]
\caption{Uncertainty estimation results $MS_{ood+ws}$ on both OOD \& WS tasks, the formula of $MS_{ood+ws}$ is described in Eq.~\ref{eq:weight_two_tasks}. The bold font annotates the best performance among a subregion. This bold font aligns with methodologies employed in similar studies on uncertainty estimation, including those detailed in Table 14 of~\citet{stadler2021graph} and Table 2 of~\citet{zhao2020uncertainty}.}
\label{tab:weighted_ner}
\centering
\scriptsize
\begin{tabular}{l|l|rrrrr|rrrrr|r}
\hline
\multicolumn{1}{c|}{\multirow{2}{*}{Data}} & \multicolumn{1}{c|}{\multirow{2}{*}{Model}} & \multicolumn{5}{c|}{weighted AUROC on both OOD \& WS task}                                                                        & \multicolumn{5}{c|}{weighted AUPR on both OOD \& WS task}                                                                         & \multicolumn{1}{l}{\multirow{2}{*}{F1}} \\
\multicolumn{1}{c|}{}                      & \multicolumn{1}{c|}{}                       & \multicolumn{1}{l}{Va.} & \multicolumn{1}{l}{Dis.} & \multicolumn{1}{l}{Al.} & \multicolumn{1}{l}{Ep.} & \multicolumn{1}{l|}{En.} & \multicolumn{1}{l}{Va.} & \multicolumn{1}{l}{Dis.} & \multicolumn{1}{l}{Al.} & \multicolumn{1}{l}{Ep.} & \multicolumn{1}{l|}{En.} & \multicolumn{1}{l}{}                                                                  \\ \hline
\multirow{5}{*}{Mov-Sim} &Dropout               & - & - & 68.66 & 71.09 & 72.11 & - & - & 27.98 & 34.06 & 31.28 & \bf83.94  \\
&PN          & \bf79.34                 & 54.41                 & 66.56 & \bf79.34 & 65.14 & 40.88                 & 16.90                 & 30.27 & 40.88 & 27.72 & 82.43  \\
&E-NER                 & 77.58                 & 59.05                 & 76.82 & 77.58 & 77.61 & 36.40                 & 20.44                 & 35.72 & 36.40 & 36.05 & 70.63  \\
&SLPN(w/o softplus)          & 60.12                 & 39.66                 & 45.35 & 60.12 & 37.33 & 28.72                 & 16.57                 & 23.47 & 28.72 & 19.36 & 66.95  \\
&Ours(SLPN) & 78.37                 & 54.22                 & 64.40 & 78.37 & 62.20 & \bf47.23                 & 16.93                 & 30.43 & \bf47.23 & 26.21 & 83.37  \\
\hline
\multirow{5}{*}{MIT-Res} &Dropout               & - & - & 61.10 & 65.86 & 63.34 & - & - & 36.66 & 47.90 & 41.15 & 74.60  \\
&PN          & 69.77                 & 66.62                 & 61.99 & 69.77 & 67.03 & 46.56                 & 39.33                 & 38.71 & 46.56 & 42.03 & 74.37  \\
&E-NER                 & 67.74                 & 67.29                 & 65.62 & 67.74 & 67.30 & 41.62                 & 40.58                 & 39.91 & 41.62 & 40.58 & 69.08  \\
&SLPN(w/o softplus)          & 50.78                 & 50.05                 & 52.48 & 50.78 & 49.92 & 32.97                 & 31.62                 & 33.43 & 32.97 & 32.37 & 62.16  \\
&Ours(SLPN) & \bf70.01                 & 49.14                 & 57.17 & \bf70.01 & 53.02 & \bf49.91                 & 32.08                 & 35.23 & \bf49.91 & 34.85 & \bf74.65  \\
\hline
\multirow{5}{*}{Mov-Com} &Dropout               & - & - & 55.82 & 56.44 & 56.13 & - & - & 17.01 & 18.68 & 17.40 & \bf72.51  \\
&PN          & 72.65                 & 68.07                 & 71.08 & 72.65 & 69.43 & 28.88                 & 22.93                 & 27.47 & 28.88 & 25.99 & 70.13  \\
&E-NER                 & 77.93                 & 73.77                 & 77.68 & 77.93 & 75.55 & 34.32                 & 25.34                 & 29.48 & 34.32 & 27.99 & 67.21  \\
&SLPN(w/o softplus)          & 60.77                 & 54.44                 & 57.91 & 60.77 & 55.56 & 25.18                 & 20.93                 & 24.32 & 25.18 & 22.71 & 66.05  \\
&Ours(SLPN) & \bf81.31                 & 48.52                 & 71.18 & \bf81.31 & 57.11 & \bf38.47                 & 17.50                 & 25.53 & \bf38.47 & 20.70 & 70.97 

 \\ \hline
\end{tabular}
\end{table*}

\begin{table}[]
\caption{Size statistics on the three cases in three datasets.}
\label{tab:three_case_statistics}
\centering
\scriptsize
\begin{tabular}{l|l|r|rrr}
\hline
   \multicolumn{1}{c|}{Data}                           &    \multicolumn{1}{c|}{Model}                  & \multicolumn{1}{c|}{$e^{ng}$} & \multicolumn{1}{c}{$e^s$} & \multicolumn{1}{c}{$\hat{e}^{p}$} & \multicolumn{1}{c}{$\hat{e}^g$} \\ \hline
\multirow{5}{*}{Mov-Sim} & Dropout               & 4412                                                    & 3055                      & 488                               & 869                             \\
                              & PN          & 4475                                                    & 2974                      & 551                               & 950                             \\
& E-NER &4847	&2665	&923	&1259 \\
                              & SLPN (w/o softplus)         & 4991                                                    & 2654                      & 1067                              & 1270                            \\
                              & Ours (SLPN) & 4426                                                    & 3060                      & 502                               & 864                             \\ \hline
\multirow{5}{*}{MIT-Res}   & Dropout               & 7217                                                    & 3793                      & 1043                              & 2381                            \\
                              & PN          & 7187                                                    & 3667                      & 1013                              & 2507                            \\
&  E-NER &7297	&3456	&1123	&2718
\\
                              & SLPN (w/o softplus)         & 7904                                                    & 3646                      & 1730                              & 2528                            \\
                              & Ours (SLPN) & 7237                                                    & 3872                      & 1063                              & 2302                            \\ \hline

\multirow{5}{*}{Mov-Com}   & Dropout               & 5551 & 3039 & 1019 & 1493 \\
    & PN         & 5772 & 3004 & 1240 & 1528 \\
&  E-NER & 5689 & 2722 & 1157 & 1810 \\
& SLPN (w/o softplus)        & 6043 & 2985 & 1511 & 1547 \\
& Ours (SLPN) & 5746 & 3045 & 1214 & 1487                       \\ \hline
\end{tabular}
\end{table}

\noindent \textbf{OOD entity construction \& data split.}
Our OOD entities are constructed using the leave-out method. Specifically, given an NER dataset with different kinds of entity labels, we count the number of entities for each label. Subsequently, we select and leave out $m$ labels with the lowest entity counts. This choice is made to ensure that there is a sufficient amount of data available for training and validation purposes.
After applying the leave-out method, we represent the remaining labels as ${S}^{in}$, which includes $c$ labels, and the corresponding text sets as $D^{in}$. Similarly, we represent the labels that were left out as ${S}^{out}$, which contains $m$ OOD labels, and the corresponding text sets as $D^{out}$. All text samples in $D^{in}$ are labeled only with entities from ${S}^{in}$ and do not include any labels from ${S}^{out}$. Conversely, all text samples in $D^{out}$ must contain at least one label from ${S}^{out}$.

We use 80\% of the samples from $D^{in}$ for training and 10\% for validation. Our testing set comprises the remaining 10\% of the samples from $D^{in}$ and all samples from $D^{out}$.

\subsubsection{Evaluation on OOD Detection}
Our uncertainty estimation is evaluated via OOD detection at the entity level (e.g., ``New York'' is an entity with the label ``LOC''). The reason for using entity-level evaluation is detailed in Sec.~\ref{sec:reason_entity_eval}.

\noindent\textbf{Wrong-span (WS) entities.} However, OOD detection evaluation in the NER task faces challenges related to wrong-span (WS) entities. Unlike traditional image or text sample-level classification, NER tasks require the prediction of entity spans first. 
An entity may span one or several tokens. There are the following three cases related to OOD detection: (1) the predicted OOD entity exactly matches a true OOD entity; (2) the predicted OOD entity partially matches a true OOD entity on some tokens; (3) the predicted OOD entity does not match a true OOD entity on any tokens. We denote the second and third cases as ``WS''.

\noindent\textbf{Three kinds of entities.} Then, because these WS entities do not have ground truth ID/OOD labels, these WS entities are inapplicable for OOD detection evaluation. Besides, we are also interested in whether our UE-NER model can handle WS entity prediction as well. As a result, we aim to evaluate our UE-NER model $\Phi$ by both OOD detection and WS entity predictions. Because the entities applicable for evaluating WS entity prediction might be inapplicable for evaluating OOD detection, we divide the ground truth entities and predicted entities into three parts: (1) Unique predicted entities $\hat{e}^{p}$, which do not exist in the ground truth and thus are the WS entities; (2) Unique ground-truth entities $\hat{e}^{g}$, which are the entities that do not appear in the predicted entities; 
(3) Shared entities $e^{s}$, which are the predicted entities matching the ground truth.

Then, all predicted entities, including shared entities, are represented as $e^p=e^s+\hat{e}^p$. Original ground-truth entities (without ``WS'' labels) are denoted as $e^{og}=e^s+\hat{e}^g$, and new ground-truth entities (including ``WS'' labels) are represented as $e^{ng}=e^s+\hat{e}^g+\hat{e}^{p}$.

\noindent\textbf{Entities applied to OOD or WS detection.} For NER OOD detection, 
the ground-truth labels in OOD detection should be binary, ``ID'' and ``OOD'' labels, while NER ground-truth labels have three: ``ID'', ``OOD'' and ``WS'' labels. As a result, we divide NER OOD detection into two subset for the evaluation. One subset has entities ($e^{og}=e^s+\hat{e}^g$) with ``ID'' and ``OOD'' for evaluating NER OOD detection, the other subset  has $e^{ng}=e^s+\hat{e}^g+\hat{e}^{p}$ entities for evaluating WS detection. For OOD detection task, we take ``OOD'' labels as 1 and ``ID'' labels as 0. For WS detection, we take ``WS'' labels as 1, ``ID'' and ``OOD'' labels as 0. We list the applied entities of these two cases in Tab.~\ref{tab:ood_intro}.

\subsubsection{Experimental Settings}
\noindent \textbf{Baselines.} Because UE-NER is underexplored, we use three baselines in our experiments: (1) Dropout~\cite{gal2016dropout}, which is an ensemble-based method to approximate BNN. It needs to run multiple times for the uncertainty estimation while our SLPN can get the estimated uncertainty by only running once. (2) PN~\cite{charpentier2020posterior}, which has been revised into token-level PN for UE-NER task, introduced in Sec.~\ref{sec:post_net}. (3) E-NER~\cite{zhang2023ner} learns importance weights via evidence distribution and adds a regularization for increasing learned uncertainty of the wrong prediction.

\noindent \textbf{Ablation Settings.} Besides PN, we design SLPN (w/o softplus) for the ablation study. The SLPN (w/o softplus) removes the softplus in Eq.~\ref{eq:QKV}. 

\noindent \textbf{Uncertainty Metrics.} We measure uncertainty estimation performance using five types of uncertainty. Specifically, Dissonance (Dis.) and vacuity (Va.) uncertainties are concepts proposed in the domain of evidential theory~\cite{sensoy2018evidential}. (1) Dissonance uncertainty refers to conflicting evidence, where the evidence for a particular class is similar to the evidence for other classes. (2) Vacuity uncertainty indicates a lack of evidence, where the evidence for all classes is of very small magnitude~\cite{lei2022uncertainty}. Besides, aleatoric (Al.) and epistemic uncertainty (Ep.) are proposed from the probabilistic view. (3) Aleatoric uncertainty arises from the inherent stochastic variability in the data generation process, such as noisy sensor data~\cite{dong2022cml}. (4) Epistemic uncertainty stems from our limited knowledge about the data distribution, like OOD data. 
Moreover, we also consider (5) uncertainty calculated by entropy. We select the best-performing metric for each method from the five available uncertainty metrics. These five types of uncertainty are all measured via AUROC and AUPR~\cite{hu2021uncertainty,zhao2020uncertainty,malinin2018predictive,hendrycks2016baseline,dong2022semi,yu2023uncertainty}. 
More details about the five uncertainty metrics are in Sec.~\ref{sec:eq_five_metrics}.

For Tables~\ref{tab:weighted_ner}, \ref{tab:ood_ner}, and \ref{tab:ws_ner}, we annotate the best performance within a subregion in bold font. This practice aligns with methodologies employed in similar studies on uncertainty estimation, including those detailed in Table 14 of~\citet{stadler2021graph} and Table 2 of~\citet{zhao2020uncertainty}.

\noindent \textbf{Performance combined OOD and WS detection performance.} Because we have OOD detection and WS detection tasks on NER uncertainty estimation, we propose to merge the results of the two tasks. This will enable us to determine which UE-NER model is better. As a result, we merge them by weighting the OOD detection results and WS detection results based on the size ratio between $e^s$ and $\hat{e}^{p}$, as shown below.
\begin{equation}
\label{eq:weight_two_tasks}
MS_{ood+ws}=\frac{e^s}{e^s+\hat{e}^{p}}MS_{ood} + \frac{\hat{e}^{p}}{e^s+\hat{e}^{p}}MS_{ws}
\end{equation}
Where $MS_{ood+ws}$ represents the metric score weighted by the respective OOD task metric score $MS_{ood}$ and the WS task metric score $MS_{ws}$.

\begin{table*}[]
\caption{Uncertainty estimation results on OOD task. The usage of bold font is the same as Table~\ref{tab:weighted_ner}. }
\label{tab:ood_ner}
\centering
\scriptsize
\begin{tabular}{l|l|rrrrr|rrrrr|r}
\hline
\multicolumn{1}{c|}{\multirow{2}{*}{Data}} & \multicolumn{1}{c|}{\multirow{2}{*}{Model}} & \multicolumn{5}{c|}{AUROC on OOD task}                                                                        & \multicolumn{5}{c|}{AUPR on  OOD task}                                                                         & \multicolumn{1}{l}{\multirow{2}{*}{F1}} \\
\multicolumn{1}{c|}{}                      & \multicolumn{1}{c|}{}                       & \multicolumn{1}{l}{Va.} & \multicolumn{1}{l}{Dis.} & \multicolumn{1}{l}{Al.} & \multicolumn{1}{l}{Ep.} & \multicolumn{1}{l|}{En.} & \multicolumn{1}{l}{Va.} & \multicolumn{1}{l}{Dis.} & \multicolumn{1}{l}{Al.} & \multicolumn{1}{l}{Ep.} & \multicolumn{1}{l|}{En.} & \multicolumn{1}{l}{}                                                                  \\ \hline
\multirow{5}{*}{Mov-Sim} &Dropout               & - & - & 69.55 & 69.67 & 72.64 & - & - & 29.61 & 34.71 & 32.62 & \bf83.94 \\
&PN          & 81.73                 & 53.41                 & 65.60 & 81.73 & 63.73 & 43.25                 & 17.47                 & 30.20 & 43.25 & 26.94 & 82.43 \\
&E-NER                 & \bf84.20                 & 60.47                 & 84.10 & \bf84.20 & 84.02 & 41.44                 & 19.92                 & 39.97 & 41.44 & 39.95 & 70.63 \\
&SLPN(w/o softplus)          & 55.37                 & 31.44                 & 36.33 & 55.37 & 26.81 & 23.33                 & 12.96                 & 15.99 & 23.33 & 12.63 & 66.95 \\
&Ours(SLPN) & 81.29                 & 53.59                 & 64.10 & 81.29 & 61.27 & \bf50.31                 & 17.56                 & 30.77 & \bf50.31 & 25.57 & 83.37 \\
\hline
\multirow{5}{*}{MIT-Res} &Dropout               & - & - & 58.01 & 64.26 & 61.08 & - & - & 39.97 & 54.36 & 45.52 & 74.60 \\
&PN          & 73.50                 & 69.44                 & 60.98 & 73.50 & 70.03 & 53.39                 & 45.16                 & 43.07 & 53.39 & 47.88 & 74.37 \\
&E-NER                 & \bf76.67                 & 75.76                 & 74.53 & \bf76.67 & 75.76 & 51.27                 & 49.79                 & 49.11 & 51.27 & 49.79 & 69.08 \\
&SLPN(w/o softplus)          & 44.30                 & 43.92                 & 46.22 & 44.30 & 41.69 & 32.66                 & 33.74                 & 33.78 & 32.66 & 31.41 & 62.16 \\
&Ours(SLPN) & 75.13                 & 45.76                 & 54.85 & 75.13 & 50.96 & \bf58.93                 & 35.63                 & 38.73 & \bf58.93 & 38.62 & \bf74.65 \\
\hline
\multirow{5}{*}{Mov-Com} &Dropout               & -                  & -                  & 50.38 & 50.75 & 50.74 & -                  & -                  & 12.33 & 14.27 & 12.52 & \bf72.51 \\
&PN          & 75.81                 & 68.86                 & 72.43 & 75.81 & 70.59 & 25.47                 & 18.85                 & 23.92 & 25.47 & 21.65 & 70.13 \\
&E-NER                 & 86.43                 & 79.90                 & 85.41 & 86.43 & 82.65 & 39.83                 & 26.54                 & 32.57 & 39.83 & 30.52 & 67.21 \\
&SLPN(w/o softplus)          & 59.59                 & 50.07                 & 53.81 & 59.59 & 50.47 & 18.71                 & 12.60                 & 16.90 & 18.71 & 13.94 & 66.05 \\
&Ours(SLPN) & \bf87.39                 & 44.28                 & 71.63 & \bf87.39 & 55.35 & \bf39.85                 & 12.47                 & 21.41 & \bf39.85 & 15.24 & 70.97
\\ \hline   
\end{tabular}
\end{table*}

\begin{table*}[]
\caption{Uncertainty estimation results on WS task. The usage of bold font is the same as Table~\ref{tab:weighted_ner}.}
\label{tab:ws_ner}
\centering
\scriptsize
\begin{tabular}{l|l|rrrrr|rrrrr|r}
\hline
\multicolumn{1}{c|}{\multirow{2}{*}{Data}} & \multicolumn{1}{c|}{\multirow{2}{*}{Model}} & \multicolumn{5}{c|}{AUROC on WS task}                                                                        & \multicolumn{5}{c|}{AUPR on  WS task}                                                                         & \multicolumn{1}{l}{\multirow{2}{*}{F1}} \\
\multicolumn{1}{c|}{}                      & \multicolumn{1}{c|}{}                       & \multicolumn{1}{l}{Va.} & \multicolumn{1}{l}{Dis.} & \multicolumn{1}{l}{Al.} & \multicolumn{1}{l}{Ep.} & \multicolumn{1}{l|}{En.} & \multicolumn{1}{l}{Va.} & \multicolumn{1}{l}{Dis.} & \multicolumn{1}{l}{Al.} & \multicolumn{1}{l}{Ep.} & \multicolumn{1}{l|}{En.} & \multicolumn{1}{l}{}                                                                  \\ \hline
\multirow{5}{*}{Mov-Sim} &Dropout               & - & - & 63.12 & \bf79.96 & 68.82 & - & - & 17.81 & 30.02 & 22.86 & \bf83.94 \\
&PN          & 66.43                 & 59.83                 & 71.76 & 66.43 & 72.74 & 28.11                 & 13.82                 & 30.66 & 28.11 & 31.91 & 82.43 \\
&E-NER                 & 58.47                 & 54.96                 & 55.81 & 58.47 & 59.11 & 21.83                 & 21.95                 & 23.44 & 21.83 & 24.80 & 70.63 \\
&SLPN(w/o softplus)          & 71.92                 & 60.10                 & 67.79 & 71.92 & 63.51 & \bf42.11                 & 25.55                 & 42.07 & \bf42.11 & 36.11 & 66.95 \\
&Ours(SLPN) & 60.60                 & 58.09                 & 66.26 & 60.60 & 67.87 & 28.43                 & 13.07                 & 28.34 & 28.43 & 30.11 & 83.37 \\
\hline
\multirow{5}{*}{MIT-Res} &Dropout               & - & - & \bf72.32 & 71.70 & 71.54 & - & - & 24.61 & 24.39 & 25.28 & 74.60 \\
&PN          & 56.25                 & 56.40                 & 65.63 & 56.25 & 56.18 & 21.84                 & 18.24                 & 22.93 & 21.84 & 20.87 & 74.37 \\
&E-NER                 & 40.25                 & 41.21                 & 38.21 & 40.25 & 41.27 & 11.94                 & 12.22                 & 11.61 & 11.94 & 12.24 & 69.08 \\
&SLPN(w/o softplus)          & 64.43                 & 62.97                 & 65.66 & 64.43 & 67.27 & 33.62                 & 27.16                 & 32.68 & 33.62 & \bf34.39 & 62.16 \\
&Ours(SLPN) & 51.34                 & 61.44                 & 65.61 & 51.34 & 60.52 & 17.04                 & 19.16                 & 22.46 & 17.04 & 21.10 & \bf74.65 \\
\hline
\multirow{5}{*}{Mov-Com} &Dropout               & -                  & -                  & 72.06 & \bf73.41 & 72.20 & -                  & -                  & 30.96 & 31.82 & 31.97 & \bf72.51 \\
&PN          & 64.98                 & 66.17                 & 67.82 & 64.98 & 66.63 & 37.13                 & 32.81                 & 36.07 & 37.13 & 36.50 & 70.13 \\
&E-NER                 & 57.92                 & 59.35                 & 59.48 & 57.92 & 58.86 & 21.37                 & 22.51                 & 22.21 & 21.37 & 22.03 & 67.21 \\
&SLPN(w/o softplus)          & 63.10                 & 63.07                 & 66.00 & 63.10 & 65.63 & 37.96                 & 37.39                 & 38.97 & 37.96 & \bf40.03 & 66.05 \\
&Ours(SLPN) & 66.05                 & 59.16                 & 70.04 & 66.05 & 61.52 & 35.00                 & 30.12                 & 35.88 & 35.00 & 34.41 & 70.97

\\ \hline   
\end{tabular}
\end{table*}

\subsection{Experimental Results}
\noindent \textbf{Our SLPN performs better than the baselines in weighted metric performance, which indicates that transmitted uncertainty from other tokens benefits the model performance.} Table~\ref{tab:weighted_ner} shows that our SLPN outperforms the baselines in weighted metric performance, except for AUROC on Movie-Simple. Specifically, our SLPN surpasses the baselines in both AUROC and AUPR on the MIT-Restaurant dataset. For instance, our SLPN improves AUPR by 2.01 points compared to dropout and 3.25 points compared to PN. On the Movie-Simple dataset, the AUPR also indicates that our SLPN performs better than other methods, with an improvement of 6.35 points compared to PN. Although the AUROC on Movie-Simple does not exceed the baselines, the difference from PN is less than 1 point. 
Plus, on the Movie-Complex dataset, our work also surpasses the baselines, such as a 3.38 points improvement over the E-NER in AUROC.
Taken together, these results demonstrate that the transmitted uncertainty from other tokens applied in SLPN benefits the model's performance.

\noindent \textbf{The entity size distribution of our SLPN is similar to that of the baselines, except E-NER.} Table~\ref{tab:three_case_statistics} shows that the entity distributions for the three types of entities are similar among dropout, PN, and our SLPN. The relatively greater number of unique predicted entities $\hat{e}^p$ and the lower number of unique ground truth entities $\hat{e}^g$ compared to dropout suggests that our SLPN primarily improves OOD detection rather than WS detection. Consequently, future research can focus on enhancing WS detection or both of these detection tasks.

Additionally, we observe that E-NER has relatively fewer shared entities $e^s$. We speculate that this could be due to E-NER not demonstrating as powerful NER classification performance as dropout, PN, and our SLPN in these three datasets.

\noindent \textbf{Our SLPN performs better than the baselines in OOD detection performance.} Table~\ref{tab:ood_ner} shows that E-NER performs better than our SLPN in Movie-Simple and MIT-Restaurant datsets, the E-NER sacrifices the NER classification performance. Among Dropout, PN and our SLPN, which have the similar high classification performance, our method performs better in OOD detection performance. For example, on the MIT-Restaurant dataset, our SLPN improves AUROC by 1.63 points compared to PN and 10.87 points compared to Dropout. However, on the Movie-Simple dataset, our SLPN has a difference of less than 1 point compared to PN, but our AUPR surpasses PN by 7.06 points.

\noindent \textbf{Our SLPN performs unsatisfactorily compared to the baselines in WS detection performance.} Although our SLPN performs very well in OOD detection, its performance in WS detection in Table~\ref{tab:ws_ner} is unsatisfactory. However, the sizes of WS entities ($\hat{e}^p$) are very similar among dropout, PN, and our SLPN on both datasets. For example, the sizes of $\hat{e}^p$ are 1043, 1013, and 1063 for dropout, PN, and our SLPN, respectively. This means our SLPN performs unsatisfactorily in WS detection.

\noindent \textbf{Our SLPN performs close or even better than the dropout in terms of the NER task performance.} From Table~\ref{tab:ws_ner}, our NER performance closely matches dropout, differing by less than 1 point in F1 scores on the Movie-Simple dataset. Notably, dropout is an ensemble-based approach known for enhancing model performance. Despite this, our SLPN achieves comparable or superior NER F1 scores, demonstrating its ability to enhance UE-NER performance while preserving the original NER model's effectiveness.

\noindent \textbf{The activation function softplus is important to make the model performs in a stable way.} 
When we remove the softplus operation (SLPN w/o softplus) and compare it with SLPN, we observe a significant performance decrease in both UE-NER and NER tasks. Table~\ref{tab:weighted_ner} indicates that NER F1 scores drop by over 10 points in both datasets, while UE-NER AUROC and AUPR scores decrease by more than 15 points. Thus, it is crucial to design the softplus operation in Eq.~\ref{eq:QKV} to ensure $\beta_{i}^{trans}$ remains positive.

\section{Conclusion}
Incorrect NER predictions incur significant penalties. We primarily focus on UE-NER, which differs from prior uncertainty estimation methods that focus on sample-level labeling. UE-NER centers on token-level sequential labeling, addressing the overlooked transmitted uncertainty from contextual tokens. We introduce SLPN to calculate uncertainty from both the token itself and contextual tokens, enhancing OOD detection in NER. Additionally, for OOD detection in NER, WS entities are not applicable. Thus, we divide the entities into two distinct subsets—one for OOD detection and the other for WS detection. Our experiments validate SLPN's effectiveness and the importance of considering uncertainty propagation in UE-NER.

\section{Ethical Considerations}
This study pioneers uncertainty estimation in sequential labeling, specifically in the context of Named Entity Recognition (NER). Additionally, we have innovatively proposed to account for uncertainty transmission, which is  ignored in sample-level classification.

Our research exclusively employs datasets that are publicly available, ensuring transparency and accessibility. Our usage of Flair and related datasets obey their MIT licenses.

\section{Limitations}
This paper introduces SLPN for uncertainty estimation in sequential labeling. However, SLPN exhibits two main limitations: First, it is based on the Posterior Network, and we plan to assess its generalization capabilities across other models. Second, our implementation of SLPN does not treat sequential labeling as a generative task, which would be meaningful to explore, especially in considering uncertainty propagation in generative tasks.

\bibliography{main}
\clearpage
\appendix

\section{Appendix}
\subsection{Model}
\subsubsection{Explanation of Softplus}
\label{sec:exp_softplus}
Since evidential learning is an evidence acquisition process, which means that every token in a training text contributes to learning an evidence matrix ($\boldsymbol{\beta}^{trans, t}$)~\cite{wang2023deep,sensoy2018evidential,amini2020deep}, we expect that $\boldsymbol{\beta}^{trans, t}$ has all elements (e.g., all tokens’ evidence in the respective class) greater than 0. 
Therefore, we expect every element of the evidential matrix ($\boldsymbol{\beta}^{trans, t}$) to be greater than 0.

Based on Eq~\ref{eq:trans_unc}, we understand that $\boldsymbol{\beta}^{trans, t}$ consists of two parts: the softmax part and $V$. If we expect $\boldsymbol{\beta}^{trans, t}$ to be greater than 0, the only potential negative case might be from $V$. Consequently, we anticipate that $V$ is greater than 0. Therefore, we choose the Softplus function, which is defined as follows:
\begin{equation}
\begin{aligned}
Softplus(x) = \log(1 + e^{x}).
\end{aligned}
\end{equation}

Considering the formula of Softplus, it is always greater than 0. In addition to ensuring $V$ is greater than 0 in Eq~\ref{eq:QKV}, we opt for Softplus as it helps prevent gradient explosion and gradient vanishing issues due to its smooth transition between the positive and negative parts of the input.

\subsection{Experiments}

\subsubsection{Criteria of Dataset Choice}
\label{sec:criteria_dataset}
We select the dataset based on two criteria: firstly, the dataset should contribute to reproducibility, and secondly, the dataset should not have an F1 score higher than 90\%. We prioritize high reproducibility because we aim for our work to be replicable by others. We do not anticipate achieving an F1 score higher than 90\%, as this would suggest that the dataset has already been thoroughly studied or that the model's uncertainty for that dataset is relatively low.

To meet the reproducibility criterion, we utilize the dataset provided by the Flair framework~\cite{akbik2019flair}. In adherence to the second criterion, we exclude CONLL\_03 dataset from consideration due to its 94\% F1 score in NER task. From the datasets listed in Flair framework~\cite{akbik2019flair}, we randomly select two domains: the restaurant domain and the movie domain. For the restaurant domain, we opt for the MIT-Restaurant dataset. In the movie domain, Flair offers both a simple movie dataset and a complex movie dataset. We are interested in investigating whether there exists a tradeoff between uncertainty scores and F1 scores in UE-NER. Consequently, we select the simple-movie dataset and the complex-movie dataset, which exhibit higher and lower NER performance, as measured by the F1 score, in UE-NER, respectively. As for the tradeoff, after excluding the impact of different domains, we do not observe a significant tradeoff between the quality of uncertainty estimation (measured by AUROC) and NER task performance (measured by F1 score) when comparing the same method’s AUROC and F1 between Mov-Sim and Mov-Com.

\subsubsection{Reason of Entity-Level Evaluation}
\label{sec:reason_entity_eval}
We choose entity-level evaluation instead of token level because it has more practical applications and is more commonly used in other NER works than token-level evaluation (e.g., ``New'' is a token with a label ``b-LOC,'' and ``York'' is a token with a label ``e-LOC''). Classifying ``New'' correctly and ``York'' incorrectly cannot lead to our desired correct entity.

\subsubsection{Metrics}
\label{sec:eq_five_metrics}
Below, we introduce the formulas used for the five metrics. Given a prediction from an EDL model, i.e., $\boldsymbol{\alpha}$, we have the total evidence $\alpha_0 = \sum_{k=1}^c\alpha_k$  (as in Eq.\ref{eq:sum_alpha}) where $c$ is the number of classes. The expected class probability is $\bar{\textbf{p}} = \frac{\boldsymbol{\alpha}}{\alpha_0}$.

From the evidential view, we have dissonance and vacuity uncertainty for EDL-based models. 
The dissonance uncertainty in EDL is calculated via Eq. 5 in~\citet{zhao2020uncertainty}.
\begin{equation}
    u^{\text{diss}} = \sum_{k=1}^c \frac{b_k \sum_{j \neq k}b_j \text{Bal}(b_j, b_k)}{\sum_{j\neq k}b_j}
\end{equation}
with $b_k = \frac{\alpha_k-1}{\alpha_0}$ and $\text{Bal}(b_j, b_k) = 1 - \frac{|b_j - b_k|}{b_j + b_k}$. It measures the uncertainty due to the conflicting evidence. 
The vacuity uncertainty in EDL is related to $\alpha_0$ in Eq.\ref{eq:sum_alpha}, which represents the total evidence,
\begin{equation}
     u^{\text{vac}} = \frac{c}{\alpha_0}
\end{equation}

From a probabilistic view, we have aleatoric uncertainty and epistemic uncertainty. The aleatoric uncertainty is calculated based on the projected or expected class probabilities,
\begin{equation}
    u^{\text{alea}} = \frac{1}{\max_k \bar{p}_k}
\end{equation}
The epistemic uncertainty is calculated based on total evidence in EDL-based models,
\begin{equation}
    u^{\text{epis}} = \frac{1}{\alpha_0}
\end{equation}
Because our vacuity uncertainty and epistemic uncertainty calculation are based on $\alpha_0$ and are similar, they have the same sample rank regarding uncertainty score.

For dropout models, where the aleatoric and epistemic uncertainty are calculated from a probabilistic view, please refer to~\citet{he2024semi,mukhoti2023deep}.

We also report the entropy as the uncertainty score, which is calculated with the expected categorical distribution. 
\begin{equation}
    u^{\text{entropy}} = \mathbb{H}(\bar{\textbf{p}})
\end{equation}

\end{document}